\documentclass[runningheads]{llncs}
\usepackage{graphicx}
\usepackage{amsmath,amssymb} % define this before the line numbering.
\usepackage{color}

\usepackage{tikz}
\usepackage{comment}
\usepackage{multirow}

\usepackage{bbm}
\usepackage{algorithm2e}
\usepackage{algorithmic}
\usepackage{enumitem}

\DeclareMathOperator*{\argmin}{arg\,min}
\DeclareMathOperator*{\argmax}{arg\,max}
\def\1{\mathbbm 1}
\def\yg{\mathcal Y_{G}}
\def\ygone{\mathcal Y_{G_1}}
\def\ygtwo{\mathcal Y_{G_2}}
\def\yf{\mathcal Y_{F}}
\def\df{\mathcal D_{F}}
\def\dg{\mathcal D_{G}}

\def\VGGFace2{VGGFace2}
\def\CASIA{CASIA-WebFace~}

\begin{document}
\pagestyle{headings}
\mainmatter

\def\ACCV20SubNumber{563}  % Insert your submission number here

%===========================================================
\title{This Person (Probably) Exists. Identity Membership Attacks Against GAN Generated Faces.}
\authorrunning{R. Webster et al.}
\author{Ryan Webster\inst{1} \and
Julien Rabin \inst{1} \and
Loic Simon \inst{1,2}\and 
Frederic Jurie \inst{1}}
\institute{University of Caen Normandie \and ENSI Caen
\\
\email{\{firstname.lastname\}@unicaen.fr}}

%******************
\newcommand{\loic}[1]{{\color{red} [L: #1]}}
\newcommand{\ryan}[1]{{\color{blue} [R: #1]}}
\newcommand{\juju}[1]{{\color{orange} [J: #1]}}
\newcommand{\fj}[1]{{\color{green} [FJ: #1]}}

\maketitle

%%%%%%%%% ABSTRACT

\begin{abstract}
Recently, generative adversarial networks (GANs) have achieved stunning realism, fooling even human observers. Indeed, the popular tongue-in-cheek website {\small \url{ http://thispersondoesnotexist.com}}, taunts users with GAN generated images that seem too real to believe. On the other hand, GANs do leak information about their training data, as evidenced by membership attacks recently demonstrated in the literature. In this work, we challenge the assumption that GAN faces really are novel creations, by constructing a successful membership attack of a new kind. Unlike previous works, our attack can accurately discern samples sharing the same identity as training samples without being the same samples. We demonstrate the interest of our attack across several popular face datasets and GAN training procedures. Notably, we show that even in the presence of significant dataset diversity, an over represented person can pose a privacy concern. 
\end{abstract}

\section{Introduction}

Among the recent advances of deep learning, Generative Adversarial Networks (GANs)\cite{goodfellow2014generative} are arguably the most burgeoning area of research.
This is evidenced by their rich applications such as ageing, inpainting, super-resolution, or attribute modification \cite{ledig2017photo,yu2018generative,webster2019detecting,choi2018stargan}. 
Less studied lines of research include measuring the statistical consistency between the generated distribution and the real one \cite{heusel2017gans,sajjadi2018assessing,simon2019revisiting}, or in theoretical analysis of the GAN learning problem \cite{arora2017generalization}.

\newlength{\malongueur}
\setlength{\malongueur}{.3\linewidth}
\begin{figure}

    \centerline{\em Samples from a GAN $G$ trained with a dataset $\mathcal{D}_G$ composed of $\yg$ distinct identities}
    \vspace{2mm}
    
    \centering
    \begin{tabular}{c c}
    \multicolumn{2}{c}{(a) $|\yg| = 220$}
    \\
    \includegraphics[height=\malongueur]{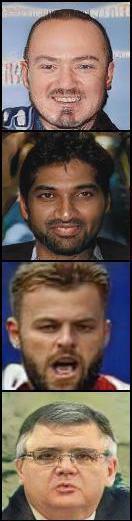}
    &
    \includegraphics[height=\malongueur]{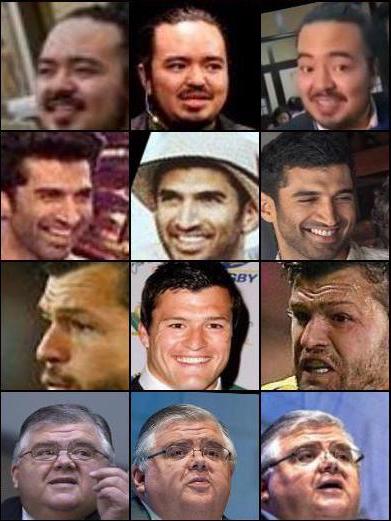}
    \\
    $G(z)$ & $x \in \mathcal{D}_G$
    %$x = G(z)$ & $y \in \mathcal{D}_G$ s.t. $Id(y) = Id(x)$
    \end{tabular}
    \hfill
    \begin{tabular}{c c}
    \multicolumn{2}{c}{(b) $|\yg| = 440$}
    \\
    \includegraphics[height=\malongueur]{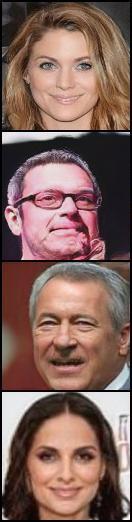}
    &
    \includegraphics[height=\malongueur]{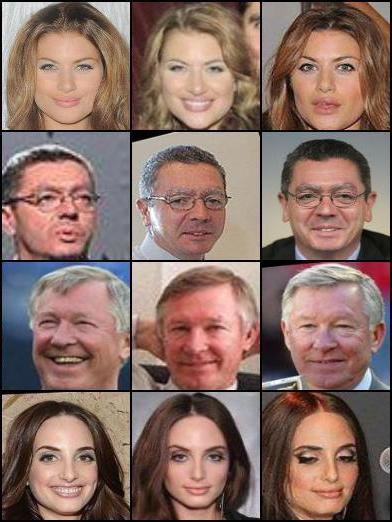}
    \\
    $G(z)$ & $x \in \mathcal{D}_G$
    \end{tabular}
    \hfill
    \begin{tabular}{c c}
    \multicolumn{2}{c}{(c) $|\yg| = 2020$}
    \\
    \includegraphics[height=\malongueur]{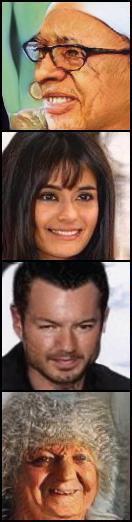}
    &
    \includegraphics[height=\malongueur]{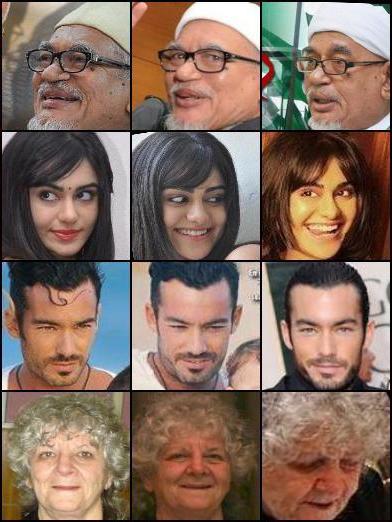}
    \\
    $G(z)$ & $x \in \mathcal{D}_G$
    \end{tabular}

    \caption{
    %\textbf{Have you already seen this person?} 
    \textbf{These people appear to exist.} 
    \em Each row displays a GAN generated image (left) with three training images (right) having the same predicted identity. 
    Images are generated with StyleGAN~\cite{karras2019style} using $N$ training images (respectively 40k, 80k and 46k) from \VGGFace2 and identified with a face identification network.
    We investigate two different scenarios in this paper: in (a) and (b) identities are evenly distributed over the datasets, where in (c) a small subset is more represented. 
    While some samples merely bear resemblance, other generated images strongly share idiosyncratic features of training identities. Such a nearest neighbor search helps factor out the ways in which GANs can generalize (via pose, lighting and expression) and elucidate overfitting on identities. This is a threat to privacy as we demonstrate in our blind identity membership attack. % (see paper).
    }
    \label{fig:teaser}
\end{figure}

It is fairly clear GANs have made incredible progress in terms of visual quality, as measured by the popular Frechet Inception Distance (FID) \cite{heusel2017gans}, or merely by sample inspection (for example, see StyleGAN samples in Fig.~\ref{fig:teaser}). However, it remains unclear in what ways GANs generalize. Several works have noted that the GAN \textit{discriminator} tends to overfit the training set \cite{arjovsky2017wasserstein,brock2018large}, in the sense that it will label hold out test images as fake. Such an observation was used to exploit the privacy of GANs in the LOGAN approach \cite{hayes2019LOGANMembershipInference}. Nonetheless, the LOGAN leak challenges mainly the discriminator, leaving the case of the generator in a somewhat gray area. In fact, several heuristics exist to show GANs do in fact produce novel data, whether it be by sample interpolation and attribute modification  as demonstrated in the first GAN works \cite{goodfellow2014generative,radford2015unsupervised}, or by the ability of GANs to generalize to novel pose and expression \cite{shen2018faceid}. Furthermore, a general belief is that the generated images are always entirely original (in terms of identity), as can be testified by the popular website
{\small \url{https://thispersondoesnotexist.com/}}.

The purpose of this article is to challenge the afore-mentioned common belief. 
We are especially interested in evaluating to what extent a GAN distribution may remain evocative of the training samples. 
Alongside the generalization aspect, this question is fundamental to evaluate the potential privacy risk posed by GANs. 
We would like to highlight that obtaining answers is becoming paramount for several reasons.
First, recent attempts to apply differential privacy mechanisms on GANs \cite{xie2018differentially,jordon2019PATEGANGeneratingSynthetic} have not been  so successful. 
Indeed, to our knowledge, published works lead to an unsatisfactory tradeoff between the privacy guarantees and the quality of the learnt distribution.
Besides, following the work of \cite{webster2019detecting} showing some form of GAN generalization, some attempts were made to leverage GANs to produce ersatz datasets aiming at more privacy. 
For instance  in \cite{webster2019generating}, a dataset, which is kept private, is used to train a GAN. Then the GAN is made publicly available for downstream tasks. 
Again, the rationale behind such a strategy is that a synthetic dataset produced from the GAN can provides the means to achieve the downstream task allegedly without compromising the privacy of the original dataset.

Notwithstanding, neural networks leak information in numerous ways about the data on which they are trained  \cite{song2017MachineLearningModels,carlini2019SecretSharerEvaluating}.
When no privacy mechanism is used, the most common way to expose potential privacy leaks from a neural network consists in applying one of several off-the-shelf attacks.
Amongst the most well known attacks are the \textit{membership inference attacks} \cite{shokri2017MembershipInferenceAttacks}.
In this type of attack, an attacker tries to discern which samples were used during model training.
In \cite{shokri2017MembershipInferenceAttacks}, training set samples could be nearly perfectly determined from model outputs on the MNIST classification task.
Such an attack typically utilizes statistics of the model output.
For example, a model that overfits will have low values of the loss function on the training samples, thus allowing for a simple thresholding attack.
Similar attacks can be seen in \cite{melis2018InferenceAttacksCollaborative,fredrikson2015ModelInversionAttacks,melis2019ExploitingUnintendedFeature}.
More recently, membership attacks have been devised against generative models \cite{liu2018PerformingCoMembershipAttacks,hayes2019LOGANMembershipInference}.
These attacks are slightly more intuitive, in the sense that generated samples leave in the same space as the data, and they may be visually similar or identical to exact training samples. In \cite{webster2019detecting}, the non-adversarial procedure of GLO \cite{bojanowski2018OptimizingLatentSpace}  reproduces some training samples verbatim, while not being able to reproduce test samples coming from the same distribution. 
Such direct leaks were not observed for typical GANs.
However, we argue that generated images can resemble training samples in even more subtle ways, as seen in Fig.~\ref{fig:teaser}: generated faces can highly resemble exact identities, even if the images themselves are quite different. 

\paragraph{\bf Identity Membership Attacks} The main contribution of this paper is to define a refined attack objective and implement such an attack. 
This new objective is meaningful in situation where a clear notion of identity exists. 
This is the case for example for datasets of face images (the central case considered in here). 
But such a notion make sense in many other contexts such as dataset of paintings (think of the identity of the painter), bio-metric data, or medical images.
In our scenario, the attacker would have access to a collection of query samples, but instead of trying to discern samples that were truly part of the training set of a GAN, the attacker should rather determine if samples with the same identity were used.
An attack of this sort shall be referred henceforth as an {\em identity membership attack}.
In addition to this fundamental difference with currently published attacks, we shall design an attack with the following properties:
\begin{itemize}
    \item contrary to LOGAN \cite{hayes2019LOGANMembershipInference}, the attacker can only exploit the generator,
    \item the attacker should not be aware of the ratio of query identities that were part of the training set (i.e, a {\em blind} attack).
\end{itemize}

We hold that if successful, such an attack would reveal as a serious hurdle for the safe exchange of GANs in sensitive contexts.
For instance, in the context of paintings or other art pieces, distributing a non-private generator might well be ruled-out for obvious copyright issues.
More importantly, consider a biometric company A releasing a generator exposing its consumer identity.  Another company B could potentially detect which of their own consumers are also clients of company A. Similar situations can pose serious issues for  medical  data,  where  revealing  a  GAN  could  breach  personal  information about a patient disease.

\begin{figure}[htb]
    \centering
    \begin{tabular}{ccc}
    % trim=left bottom right top, clip
    \includegraphics[width=.285\linewidth]{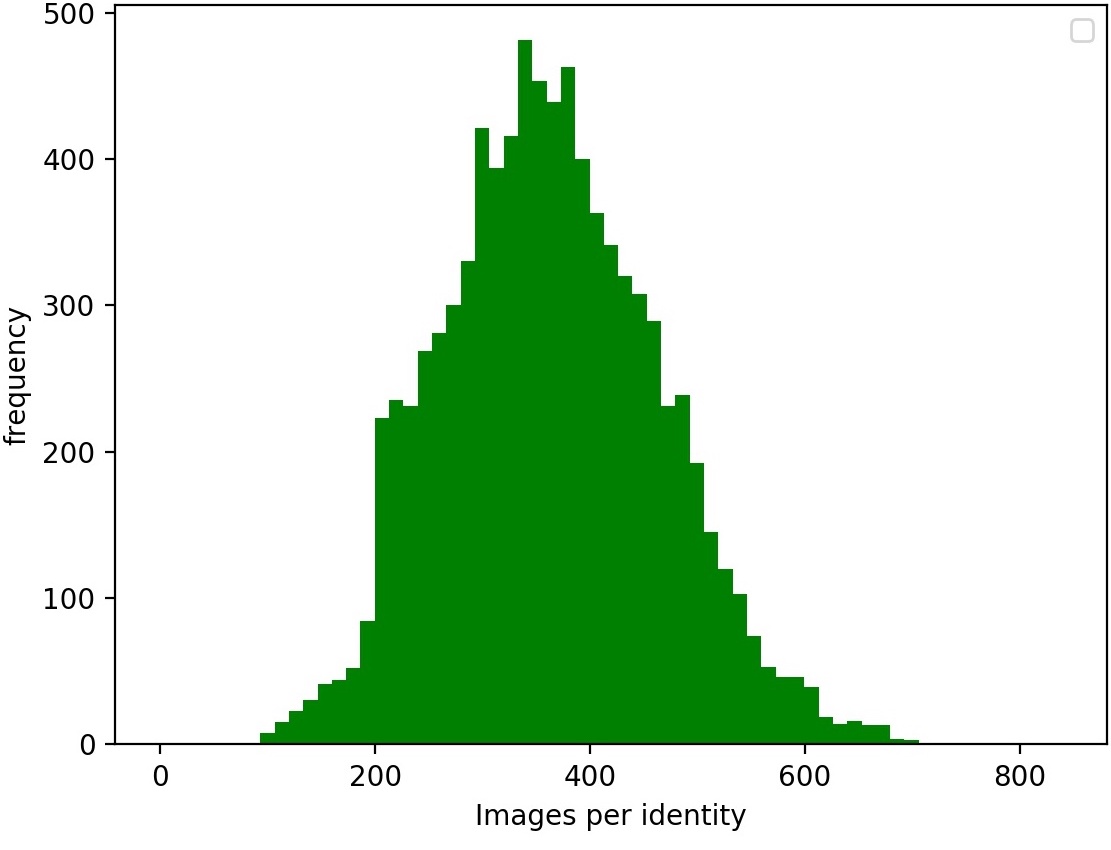}
    & \includegraphics[width=.29\linewidth]{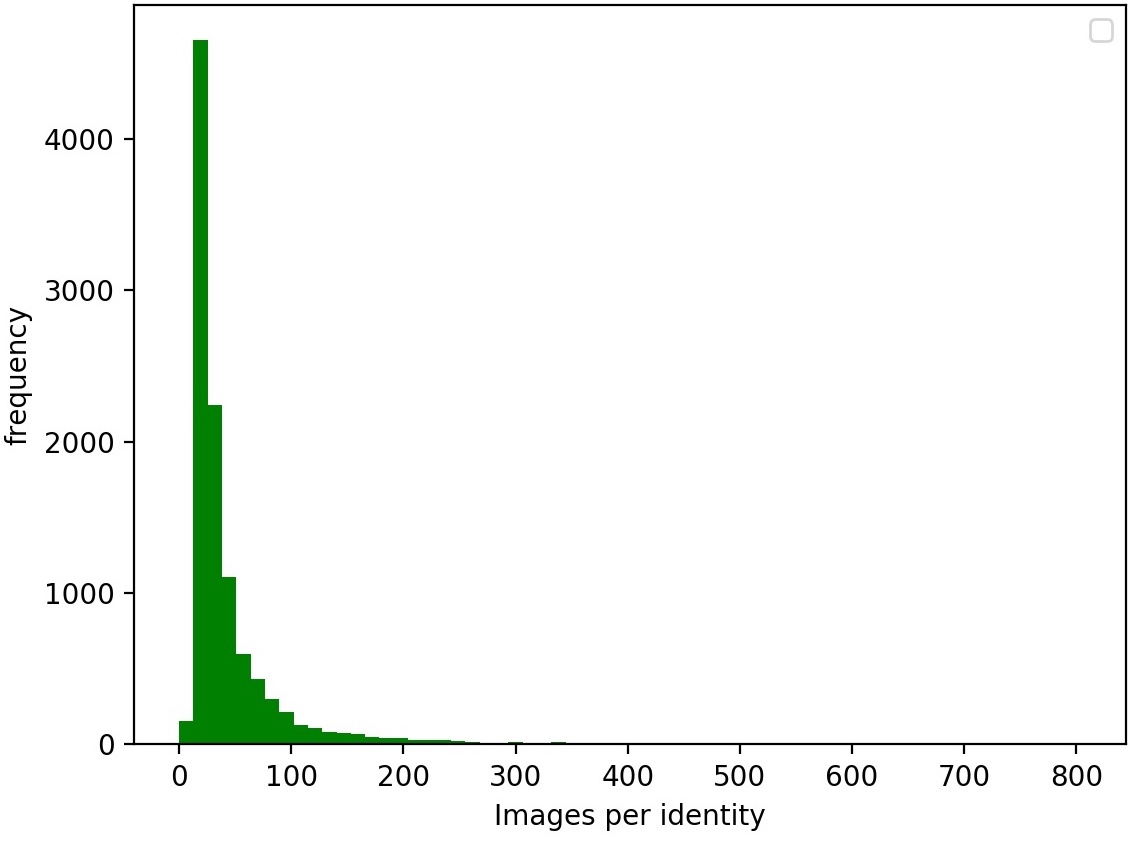}
    & \includegraphics[width=.33\linewidth, trim=0 0 0 110, clip]{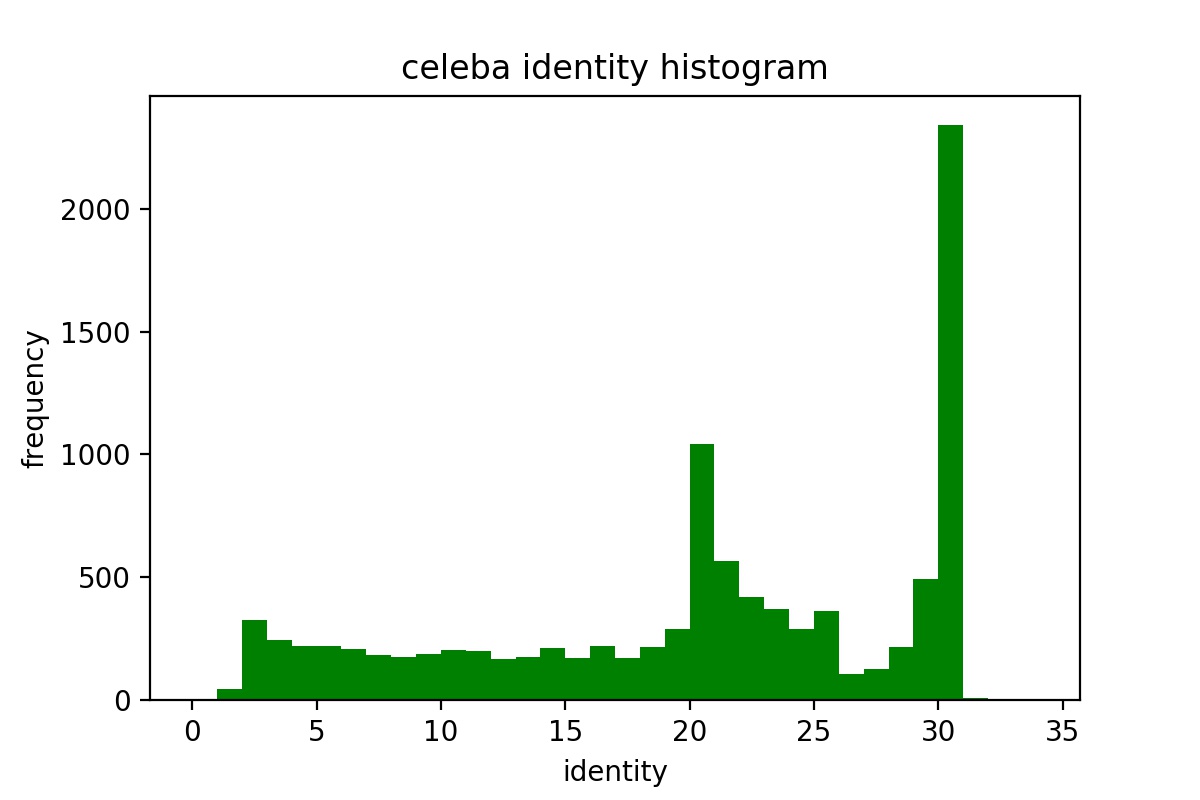}  \\
         
    \vspace*{1mm}
    \VGGFace2 &  \CASIA & CelebA
    \end{tabular}
    
    \caption{\textbf{Histogram of number of instances per identity for different datasets (\VGGFace2, \CASIA \& CelebA).} 
    \em Far from being uniform, face recognition datasets include bias in terms of the number of examples per identitiy. VGGFace2 was designed to be balanced in this sense, compared here to the \CASIA containing bias in some samples.
    }
    \label{fig:histogram_identity_datasets}
\end{figure}

\paragraph{\bf Face Generation}
In this work, we focus on GANs trained on facial image data.
Face image datasets are amongst the most popular for demonstrating GAN efficacy, partly because the human eye is particularly attuned to detect artifacts in these images and due to the wide availability of high quality face datasets \cite{kemelmacher2016megaface,cao2018vggface2,liu2018large}. Face datasets include some form of bias in regard to the real distributions of faces.
For instance, datasets are frequently composed of a decent number of instances of the same person (see Fig. \ref{fig:histogram_identity_datasets}).
Besides in some datasets \cite{wang2018deep}, the number of instances per identity is highly versatile (e.g. in MEGAFACE \cite{kemelmacher2016megaface} it ranges from 3 to 2k). One should also note that datasets dedicated to generation such as FFHQ \cite{karras2019style} may well display similar biases due to the lack of any specific safeguard.

\paragraph{\bf Outline}
The rest of the paper is organized as follows.
In Section~\ref{sec:protocol} we expose the proposed protocol used to train the face detection network, to perform the attack and to evaluate its performance.
This protocol is tested in Section~\ref{sec:attack} on various datasets (\CASIA and \VGGFace2) with different GANs (StyleGAN and LSGAN) across different attack scenarios.
Performance curves and visual evidence are provided to show membership attacks can be highly successful against GANs even in the blind scenario, and when diverse training data is used.
A discussion and perspectives on future work are given in Section~\ref{sec:conclusion}.
Reviewers will also find additional results in the supplementary material.

\section{Identity Membership Attack}
\label{sec:protocol}
As argued earlier, because standard datasets gather several instances of some individuals, training generative models on such datasets exposes those individuals to privacy leaks.
Figure \ref{fig:teaser} illustrates such a leak for a GAN trained on a subset of \VGGFace2.
There is no doubt that in many cases, the displayed generative sample is but an instance of the training identity shown under various poses (last three columns).
As such the generated samples can be leveraged to extract information on the identities that were seen during the training of the GAN.

\begin{figure}[htb]
% old editable source : https://drive.google.com/file/d/1otiAqywljIZ39ZI7iHEPAq_ty8CfJaE_/view?usp=sharing 
% new one
% https://drive.google.com/file/d/1WDZJIf9SoAEWxcp6n6zpyCRm850PKTR2/view?usp=sharing
    \centering
    %\begin{minipage}[p]{.5\linewidth}
    %\includegraphics[width=\linewidth]{figures/protocol_ECCV20.pdf}
    %\end{minipage}%
    \begin{minipage}[p]{\linewidth}
    \includegraphics[width=\linewidth]{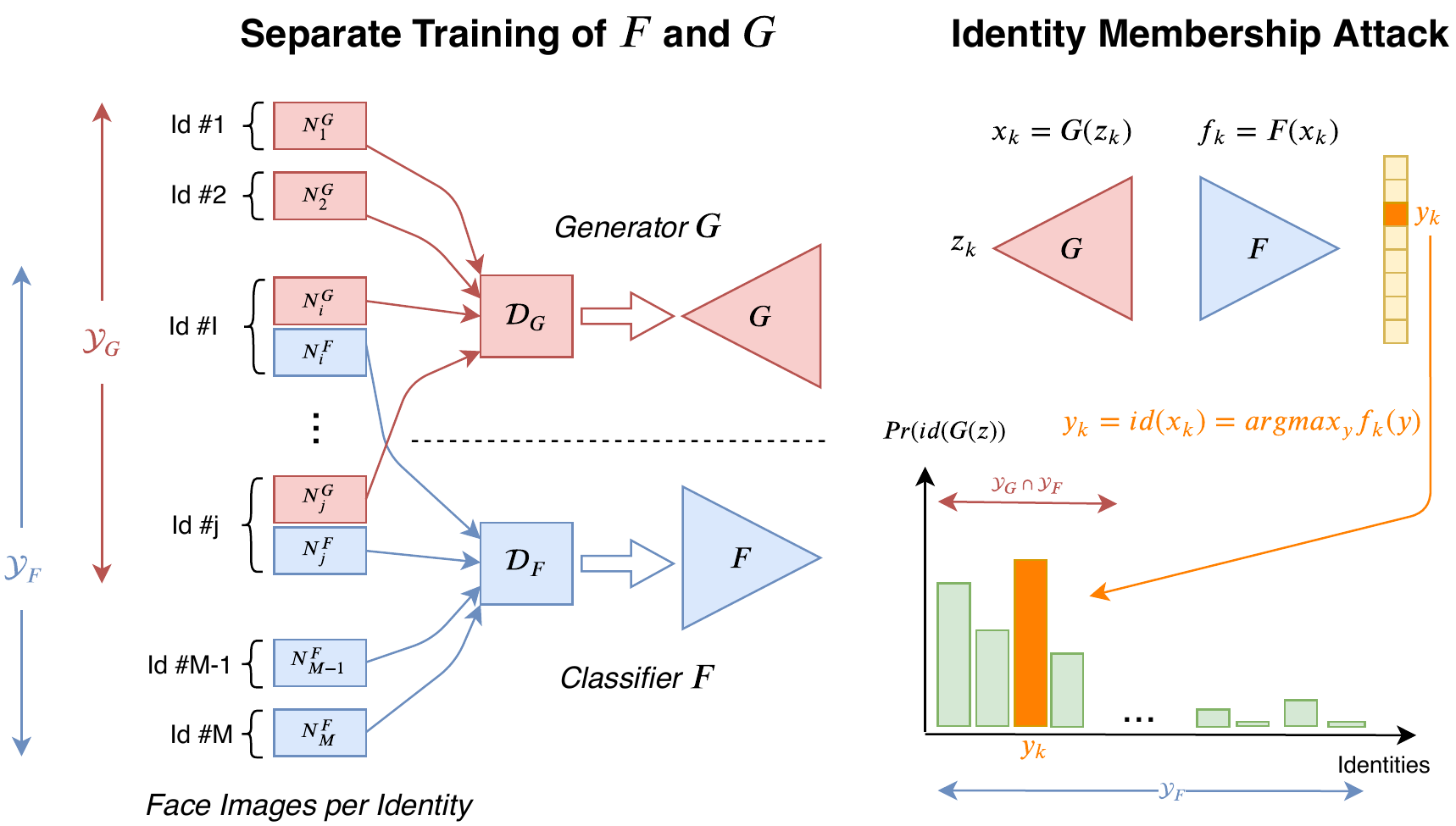}
    \end{minipage}
    \caption{Illustration of the protocol used for the attack.  A generator $G$ is trained from a dataset $\dg$ gathering image instances of identities $y\in\yg$ (red samples).
   A face classifier $F$ is trained by an attacker to recognize identities $y\in\yf$ from a separate dataset $\df$ (blue samples).
   Although the samples are completely separate, the two datasets share some common identities $\yg\cap\yf\neq\emptyset$.
   The distribution of the identities of the faces generated by $G$, according to $F$, can be used to detect overfitting in $G$, and thus to infer identities seen during the training of the generator.}
    \label{fig:protocol}
\end{figure}

% \paragraph{\bf Attack Assumptions}
\subsection{Attack Assumptions}\label{sec:Assumptions}
We consider the following attack scenario : the attacker wants to determine if instances of certain query individuals were used in the training of the GAN.
We reference this type of attack as an {\em identity membership} attack.
More formally, our  attack protocol is illustrated in Fig.~\ref{fig:protocol} and delineated as follows:
\begin{itemize}\label{sec:attack principles}
    \item A GAN $G$ is trained on a dataset $\dg:=\{(x^G_i,y^G_i) \}$ where $\forall i, y^G_i\in \yg$;
    \item The attacker trains a face identification network, $F$ on a dataset $\df=\{(x^F_i,y^F_i) \}$ where $\forall i,y^F_i\in \yf$ to recognize instances from identities $y\in\yf$;
    \item It is assumed that $\yg\cap\yf\neq\emptyset$\footnote{In practice, we shall consider a worse case situation where $\yg\subsetneq\yf$.} but $\df\cap \dg=\emptyset$  . 
    The first assumption reflects the fact that the attacker has some founded suspicion that some identities were used in the training of $G$.
    Nonetheless the second assumption ensures that no instance of $D_G$ is explicitely known to the attacker.
\end{itemize}

\subsection{Attack Algorithm}\label{sec:algo}
The basic principle of the attack consists in randomly generating faces $x_k:=G(z_k)$ for $k\in\{1,\ldots,N\}$. 
Then using the network $F$, these random samples are identified to identities $y_k:=id(x_k)$ where $id(x):=\argmax_{y\in\yf} F_y(x)$.
Eventually, the attacker will suspect the identities of $\yf$ that are more often predicted as the ones that were seen during the training of $G$.
This mechanism is described in more details in Alg.~\ref{algo:attack} and merely corresponds to thresholding the number of times the query identity $y$ was predicted by $F$ on the generated samples $x_k=G(z_k)$.

To be effective, such an algorithm requires to generate a large enough set of samples $K$.  
Typically this parameter shall be set to many times the number of identities in $\df$ i.e. $K=\lambda \times |\yf|$. 
In such case, it is natural to fix the frequency threshold $T$ to $\lambda$. We will denote this natural value of the threshold $T_{0}=\lambda$.
Yet, of course, one can trade recall for precision by increasing this threshold. In our experiments, we shall consider also $T_{1}=10\lambda$.

%\begin{figure}[ht]
  \begin{center}
  \begin{minipage}{.7\linewidth}
    \begin{algorithm}[H]
     \SetKwInput{KwData}{Inputs}
     \SetKwInput{KwResult}{Output}
     \SetAlgoLined\DontPrintSemicolon
     \SetKwFunction{algo}{identityAttack}
     \SetKwFunction{sample}{randomSample}
     \KwData{
        the query identity $y\in\yf$,\newline
        the number of generated samples $K$,\newline
        the frequency threshold $T$
     }
     \KwResult{a boolean prediction of $\1_{y\in\yg}$}
     \SetKwProg{myalg}{Algorithm}{}{}
     \SetKwProg{myproc}{Procedure}{}{}
     \myalg{\algo{$y$}}{
     \nl $k_y=0$\;
     \nl \For{$k=0$ to $K$}{
            \nl $z = \sample()$ \;
            \nl $x = G(z)$ \;
            \nl \If{$id(x)=y$} {
                \nl $k_y+=1$ \;
            }
        }
     \nl \KwRet $ \1_{k_y\geq T}$ \; % C(y) =
     }
     \;
     \caption{\em Proposed identity membership attack}
     \label{algo:attack}
    \end{algorithm}
  \end{minipage}
  \end{center}
%\end{figure}

%\paragraph{\bf Training of F}
\subsection{Classifier Training}\label{sec:training_F}
To train the classifier $F$, we use pre-trained features from the VGGFace network \cite{parkhi2015deep}.
Note that VGGFace and VGGFace2 share 53 identities. 
Therefore, to  ensure that $\df\cap \dg=\emptyset$, \emph{i.e.} $F$ is not trained on any of the sample samples as $G$, we remove these 53 identities from the VGGFace2 dataset.
Then, we pool the relu5\_3 layer of VGGFace to a $2 \times 2$ spatial resolution.
Finally, we train a single fully connected layer with $|\yf|$ output classes.
Note that $|\yf| = 8631$ for VGGFace2 and $|\yf| = 1292$ for CASIA-Webface.
Finally, models are trained in pytorch with SGD, a learning rate of .1, and a momentum .9.
This simple model for $F$ achieves an average top-1 classification accuracy of 86.1\% on VGGFace2 and 94.7\% on CASIA-Webface.
Likely, using even better classifiers should improve the membership attack accuracy, but we leave this direction for future work. 

%\paragraph{\bf Attack Evaluation}
\subsection{Attack Evaluation}\label{sec:attack_evaluation}
The attack can be seen as a binary classification problem where the attacker must classify identities from $\yf$ 
according to whether they also belong to $\yg$ or not.
It is therefore natural to evaluate the performance of  Alg.~\ref{algo:attack} in terms of precision and recall. 
Taking the attacker point of view, we will tag as positive all the identities in $\yg\cap\yf$ and the remaining ones as negative.
Therefore, denoting $C(y):=\1_{k_y\geq T}$ the decision made by the attacker, the precision and recall will be computed as follows:

\begin{equation}\label{eq:precision-recall}
    \alpha:= \frac{\sum_{y\in\yg\cap\yf} C(y) }{\sum_{y\in\yf} C(y) }
    \quad \text{ and } \quad
    \beta:=\frac{\sum_{y\in\yg\cap\yf} C(y)}{|\yg\cap\yf|}
\end{equation}

As the goal of privacy is to protect all individuals equally, an attack with high precision (and positive be it small recall) should be considered more troublesome than one with high recall. In other words, if an attack can accurately discern training information even on a small subset of data, it still violates the fundamental goal of privacy. To summarize precision recall in the following section, we'll highlight attacks with the highest $F_{1}$ score, which is the harmonic average of precision and recall as follows
\begin{equation}\label{eq:f1 score}
    F_{1}(\alpha,\beta):= 2 \cdot \frac{\alpha \cdot \beta}{\alpha + \beta }
\end{equation}

%\paragraph{\bf Visual evaluation}
\subsection{Visual Evaluation}\label{sec:visual_evaluation}
As a visual sanity check, we include a nearest neighbor search utilizing the recognition network $F$. An image or feature space nearest neighbor search is a common procedure in the GAN literature, to dispel suspicions of overfitting \cite{goodfellow2014generative,brock2018large,karras2019style}. We visually compare generated samples falling inside the GAN training set $y \in \yg$, and those detected to lie outside $y \not\in \yg$. More formally, we generate an image $x:=G(z)$, find the identity using $F$ as $y:=id(x)$ and finally perform the intra-identity nearest neighbor search as
\begin{equation}\label{eq:NN_id}
x_\text{NN} = \argmin_{x_{y_i}} {\| f(x_{y_i}) - f(x)\|}^{2}_{2}
\end{equation}
where $x_\text{NN}$ is the retrieved nearest neighbor, $f$ is the before softmax feature representation of $F$ and finally $x_{y_i}$ represents the i-th image among the instances of identity $y$. Results are displayed for three attack scenarios in Fig.~\ref{fig:teaser}. Indeed, in both the scenarios of low diversity and dataset bias, generated images highly resemble those in the training set. Intuitively, these also correspond with attack scenarios that are highly accurate at guessing training identities (see Table~\ref{table:membership attacks precision-recall \VGGFace2 and Casia} and Table ~\ref{table:membership attacks precision \VGGFace2 setting2}). Note that this visual demonstration should be robust to any distortions or variations found in generated images, as $F$ is specifically designed to be robust to natural variations in the dataset like pose or expression.

%\clearpage

\section{Identity Membership Attack Evaluation} \label{sec:attack}

\subsection{Experimental Protocol}
We evaluate the efficacy of our identity membership attack on a variety of training settings, in order to elucidate what factors in training data influence privacy. As is shown in Figure~\ref{fig:histogram_identity_datasets}, face datasets exhibit varying number of identities, and importantly bias in the number of sample they have per identity. VGGFace2 contains a relatively balanced number of samples per identity, while VGGFace2 has far more average samples per identity than CelebA for example. On the other hand, in \CASIA the number of samples per identity ranges from a few to many. In the following training settings, we denote \textit{diversity} as number of training identities and \textit{(imbalance) bias} when some identities have many more samples than others. As both $G$ and $F$ require some minimum amount of samples to train, and with our disjointness assumption for blind attacks (see Sec.~\ref{sec:Assumptions}), we also curate \CASIA to contain at least 80 samples per identity. We then explore two real world training scenarios in the following sections.

Throughout experiments, the parameter $\lambda$ discussed in Sec.~\ref{sec:algo} used to sample the Generator during attack is fixed to $\lambda=2$.

\subsection{Setting 1: Low bias and varying diversity}
\label{sec:singlesplit}%\S~\ref{sec:singlesplit}

In this setting, we simply vary the number of identities $|\mathcal Y_G|$ ranging from $30$ to $880$, taking a large set of samples from the the first $|\mathcal Y_F|$ identities.
The number of samples per identity $N^G_i$ used during the training of the generator is fixed to simulate datasets that are evenly distributed.
In addition, we consider two successful GAN methods, the Least Squares GAN (LSGAN) \cite{mao2017least} and the state of the art StyleGAN \cite{karras2019style} network. 

\begin{figure}[!htb]
    \centering
    \begin{tabular}{ccc}
        \raisebox{20mm}{\rotatebox[origin=c]{90}{\bf \VGGFace2}}
        %\hspace*{2mm}
        & \includegraphics[width = .48\linewidth,trim={.4cm 0 1.5cm 0},clip]{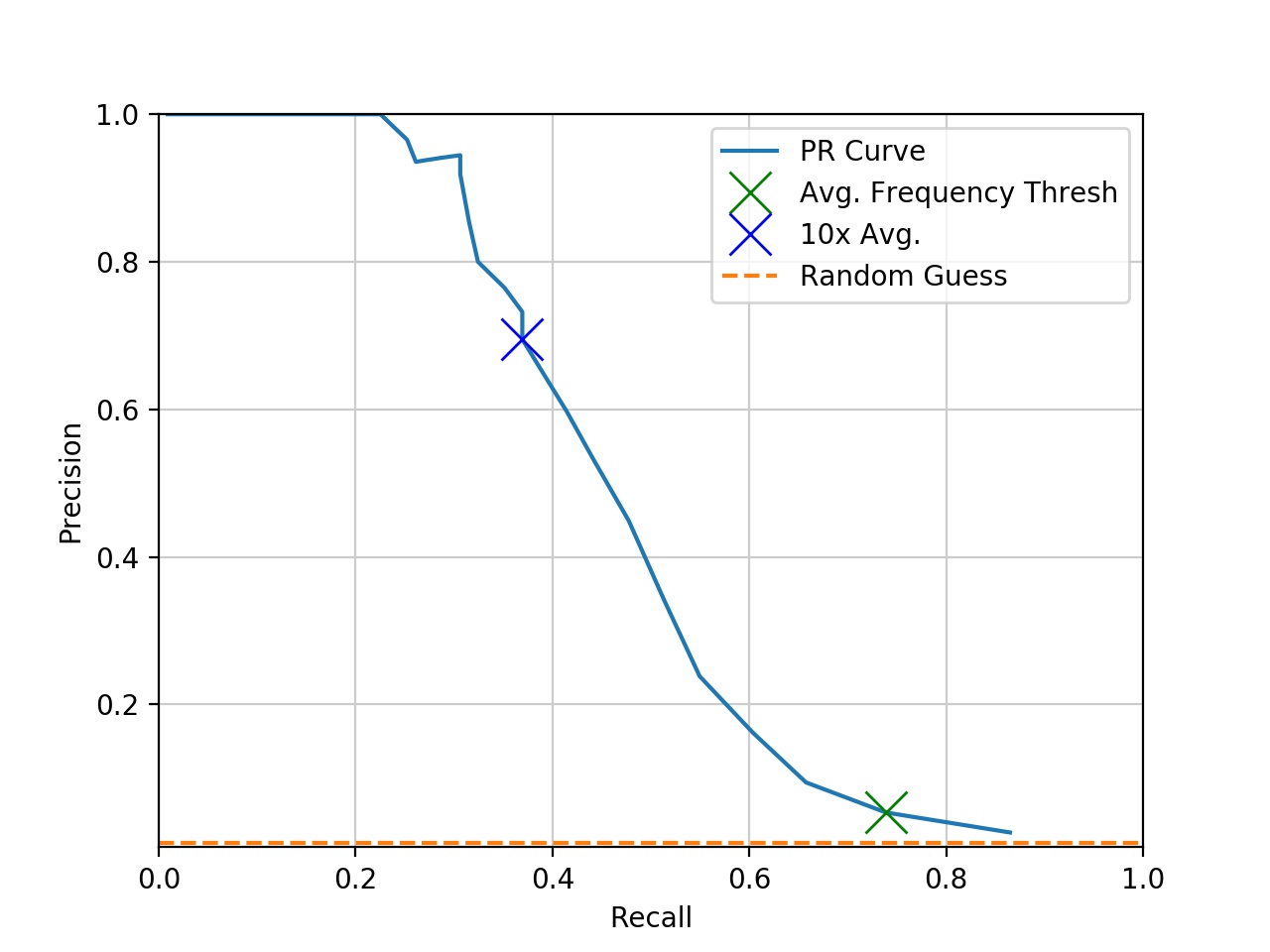}
        & \includegraphics[width = .48\linewidth,trim={.4cm 0 1.5cm 0},clip]{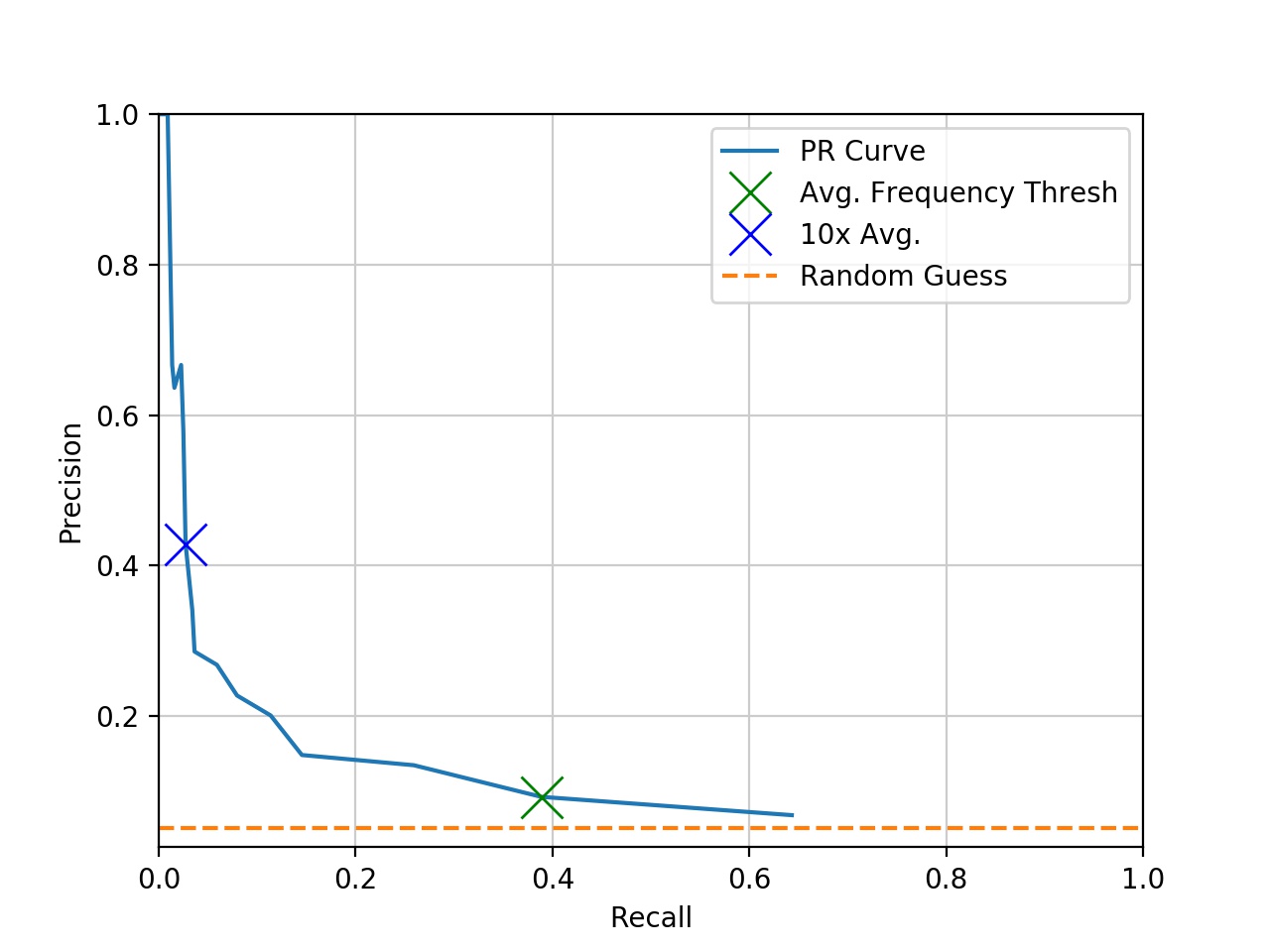}
        \\
        & $| \yg| = 111$ 
        & $| \yg| = 440$
        \\
    \end{tabular}
    
    %\vspace{1mm}
    
    \centering
    \begin{tabular}{ccc}
        \raisebox{20mm}{\rotatebox[origin=c]{90}{\bf \CASIA}}
        %\hspace*{2mm}
        &\includegraphics[width = .48\linewidth,trim={.4cm 0 1.5cm 0},clip]{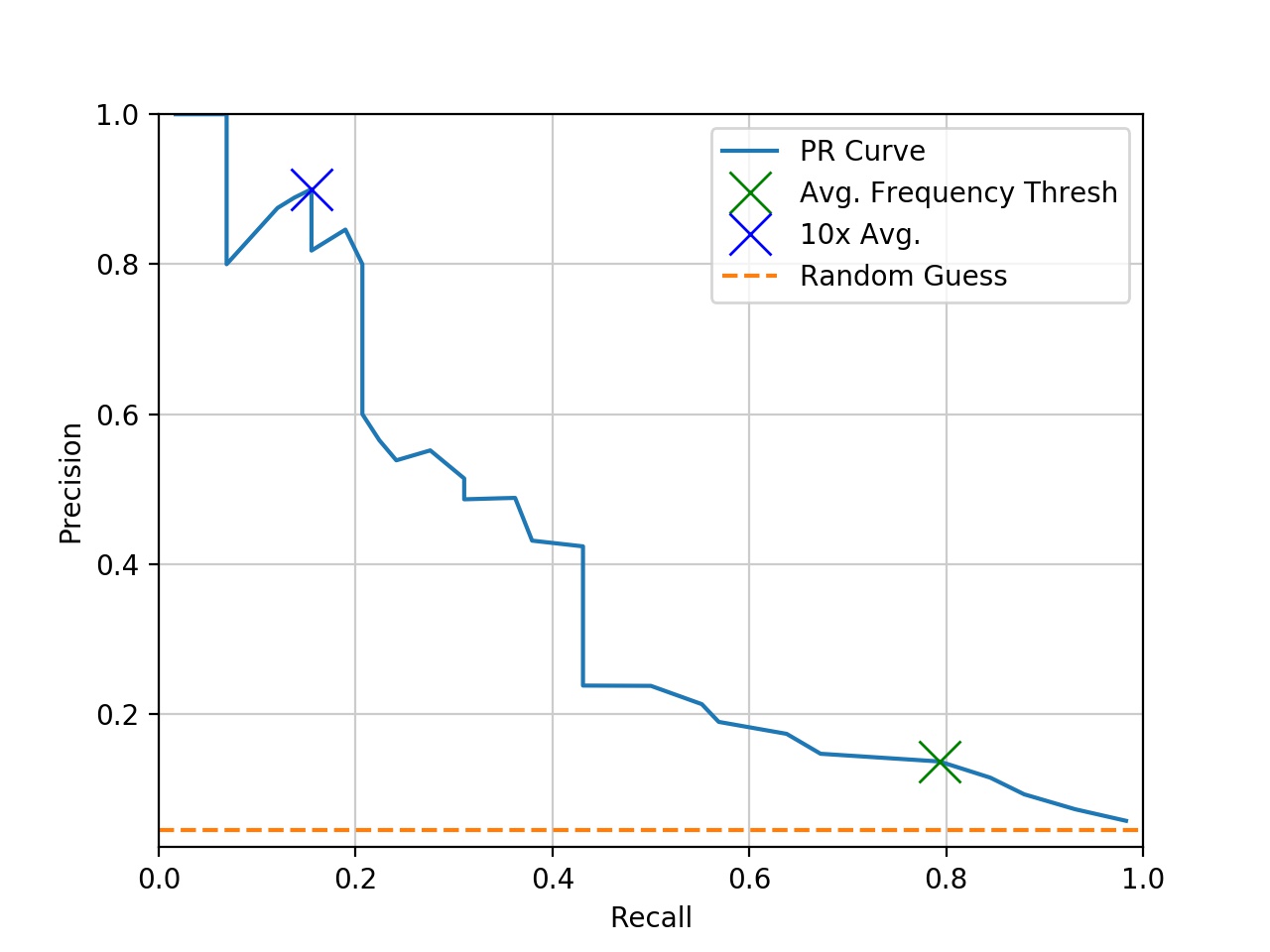}
        &\includegraphics[width = .48\linewidth,trim={.4cm 0 1.5cm 0},clip]{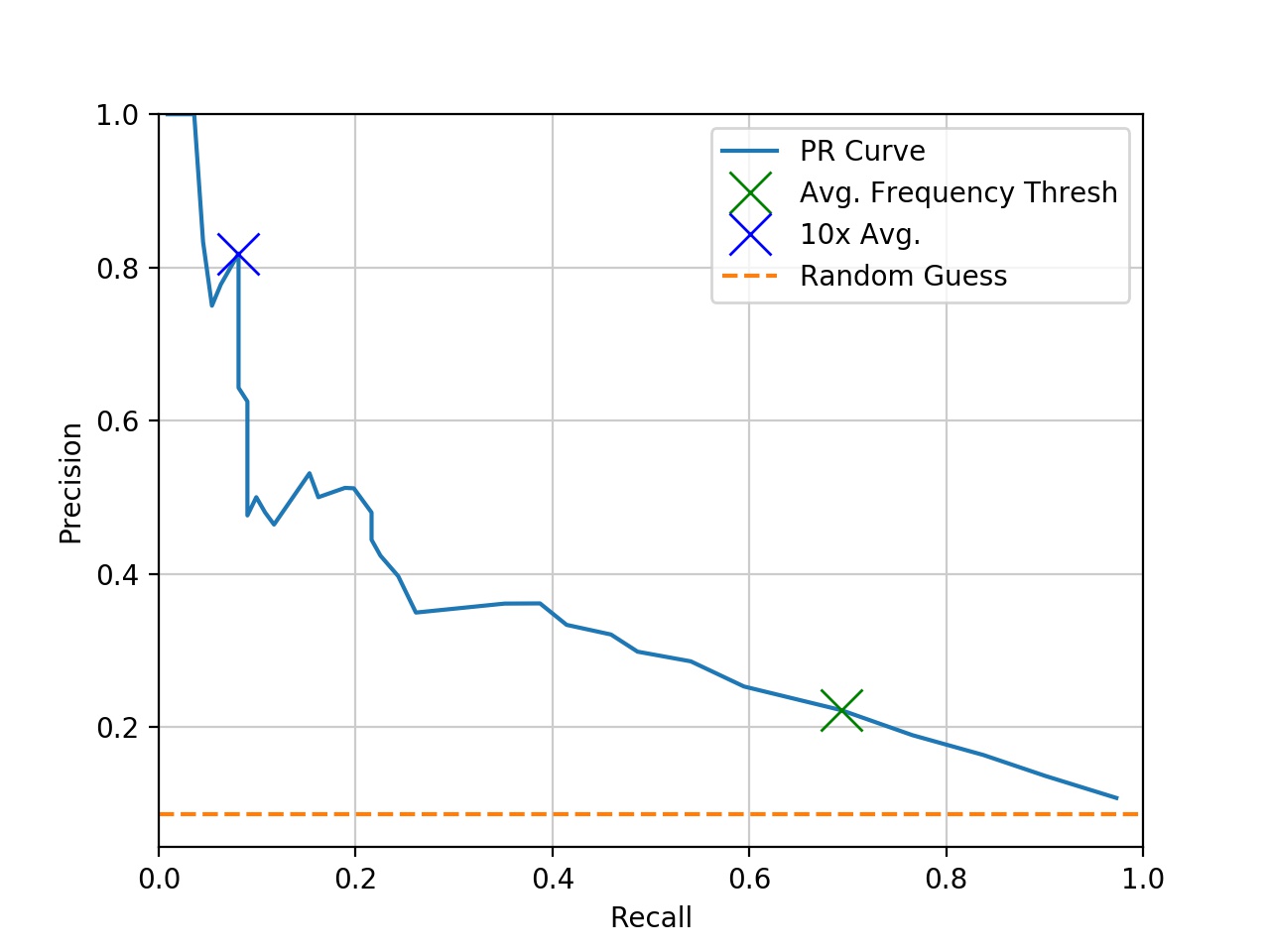}
        \\
        & $|\yg| = 58$ 
        & $|\yg| = 111$
        \\
    \end{tabular}

    \caption{\textbf{Precision-recall curves for identity membership attacks for setting 1}.
     \em Curves are obtained using Eq.\eqref{eq:precision-recall}, the Style-GAN generator, two datasets and varying diversity $\yg$. Models trained on more diverse data appear to be more private. In addition, the  threshold $T_1=10 T_0$ can discern training identities with a high precision.
    }
    \label{fig:PR_curve_diversity}
\end{figure}

Table~\ref{table:membership attacks precision-recall \VGGFace2 and Casia} gives the precision and recall rate for each setting, on two different datasets (\VGGFace2 and \CASIA).
The total number of samples used during training is indicated by $N = \sum_i N_i^G$, and is increasing with the number of identities.
Each column corresponds to a different generation training set, with a varying number of identities while the total number of identities is kept fixed ($| \yf| \approx 8$K).
$F_{1}$ scores are used to highlight the most successful attacks.
The baseline corresponding to random face identification has a precision of $\frac{|\yg|}{|\yf|}$ at any recall (witch is set to 100\% to compute $F_1$ scores.
As mentioned earlier, two different frequency thresholds $T_0$ and $T_1$ are proposed to perform the attack, $T_0$ being more conservative towards recall, while $T_1$ is more conservative with respect to precision.
Figure \ref{fig:PR_curve_diversity} displays Precision-Recall curves obtained for some settings when spanning the frequency threshold.

\subsection{Setting 2: Varying bias and high diversity}
\label{sec:doublesplit}
We also examine scenario when training data is highly diverse, yet due to natural bias in the data collection procedure or data availability, certain identities contain many more images than others. For this setting, we vary the number of biased identities but enrich each dataset with a large set of diverse unbiased faces. More formally, we take $\yg:=\ygone\cup \ygtwo$, with $\ygone$ containing many more samples per identity than $\ygtwo$. Furthermore, we consider when $|\ygtwo| \gg |\ygone|$, \emph{i.e.} the dataset has high class diversity. Such is naturally the case in \CASIA, see Fig.~\ref{fig:histogram_identity_datasets}.
One may however suspect that the GAN be prone to reproduce mainly identities from $\ygone$ (since they are more present in the dataset).
We therefore only compute the precision/recall in Eq.~\eqref{eq:precision-recall} and $F_{1}$ values in Eq.~\eqref{eq:f1 score} for the biased set $\ygone$ in lieu of $\yg$.

\begin{figure}[!htb]
    %\centerline{\em  Attack Precision-Recall curve for setting 2 (\S~\ref{sec:doublesplit}): varying bias with high diversity}
    %\vspace{2mm}
    
    \centering
    \begin{tabular}{ccc}
        \raisebox{20mm}{\rotatebox[origin=c]{90}{\bf \VGGFace2}}
        %\hspace*{2mm}
        &\includegraphics[width = .48\linewidth,trim={.4cm 0 1.5cm 0},clip]{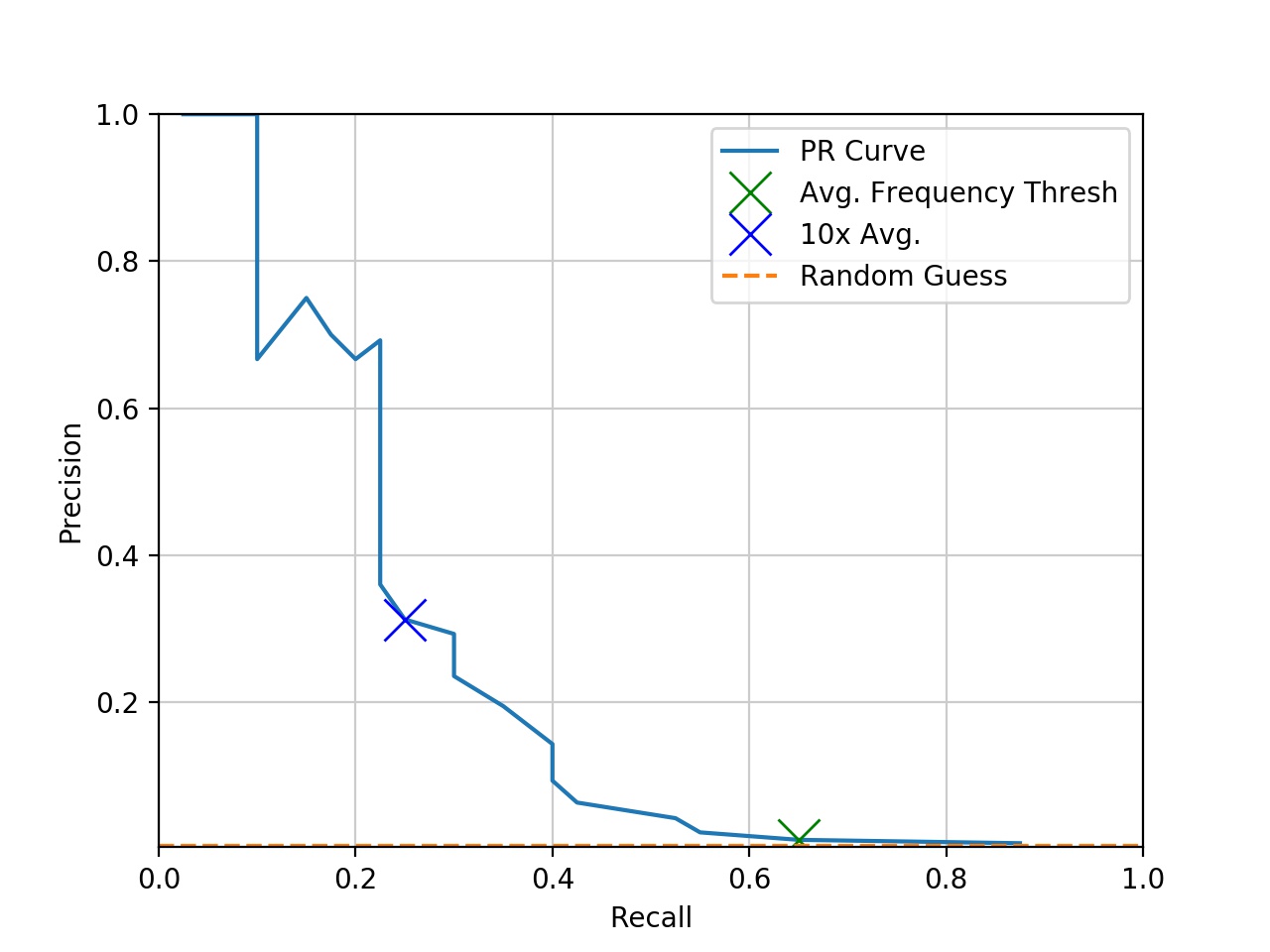}
        &\includegraphics[width = .48\linewidth,trim={.4cm 0 1.5cm 0},clip]{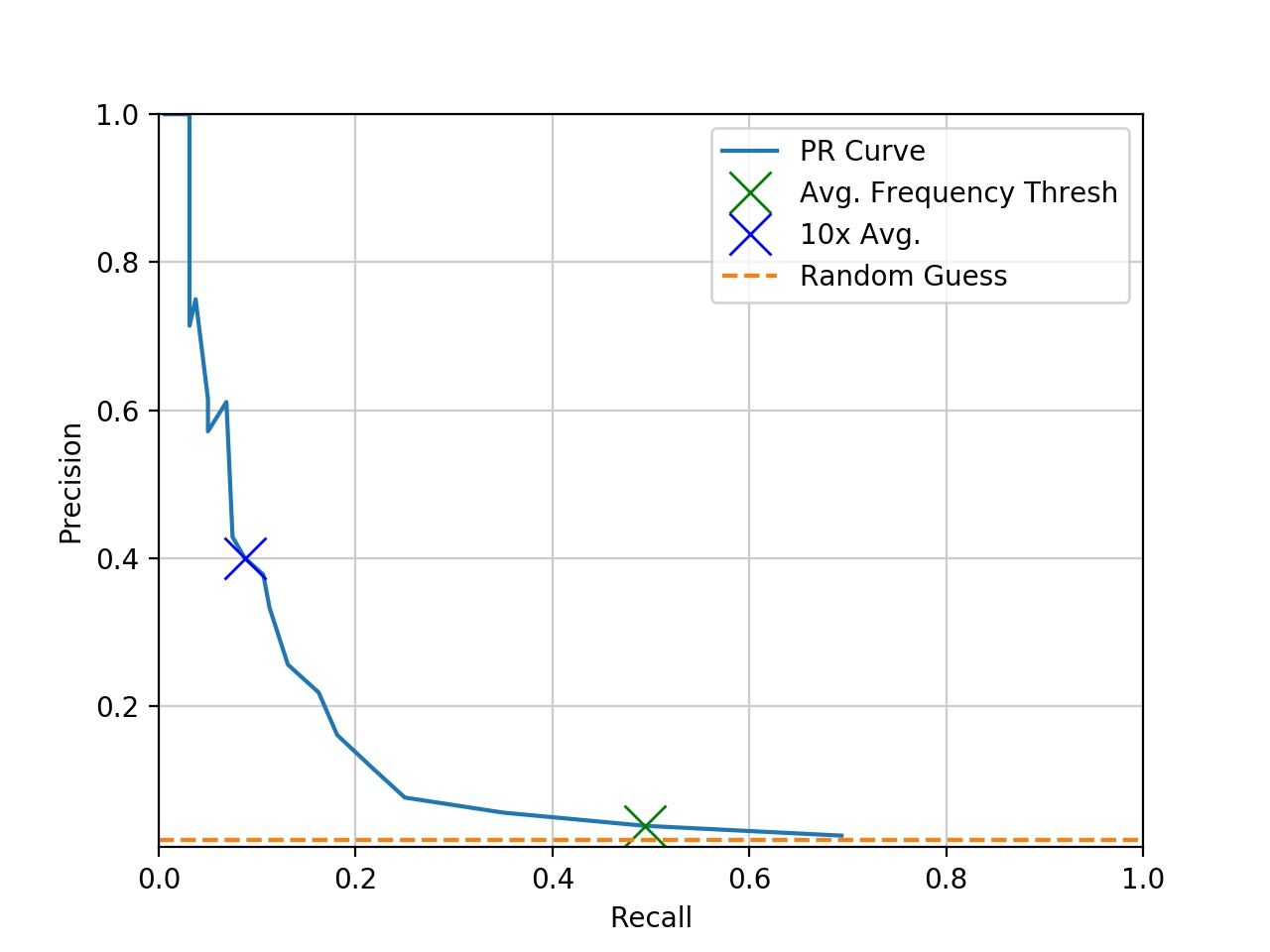}
        \\
        & $|\ygone| = 40$, $|\ygtwo| = 2000$, $N=52k$
        & $| \ygone| = 160$, $| \ygtwo| = 2000$, $N=88k$
        \\
    \end{tabular}
    
    \caption{\textbf{Precision-recall curves for membership attack on biased data described in Sec.~\ref{sec:doublesplit}.}
     \em Curves are obtained using Eq.\eqref{eq:precision-recall}, the Style-GAN generator, two datasets and varying diversity of the biased dataset $\ygone$. Even though both settings are trained with significant diversity (more than 2k identities and 50k images), the biased data in $\ygone$ is still detectable.
    }
    \label{fig:PR_curve_biased}
\end{figure}

Table~\ref{table:membership attacks precision \VGGFace2 setting2} gives the precision and recall rates of the attack in such setting on \VGGFace2 dataset.
Attack are performed on LSGAN and StyleGAN trained with an increasing number of identities for subset $\ygone$, while the second set $\ygtwo$ is fixed to 40'000 samples evenly distributed across 2000 identities.
The bias is therefore decreasing, as measured by the ratio $\frac{|\ygone|}{|\yf|}$ corresponding to the precision of random identification, as reported in the Table.
The total number of samples used during training is indicated by $N = N_1+N_2$, and is increasing with the number of identities.
Figure \ref{fig:PR_curve_biased} displays Precision-Recall curves obtained for some settings when spanning the frequency threshold.

\begin{table}[htb] %[!p]% !htb
\footnotesize %\small
\centering
\hspace*{-6mm}
\rotatebox[origin=c]{90}{\bf \VGGFace2}
\renewcommand{\arraystretch}{1.5}\setlength{\tabcolsep}{2pt}
%\hspace*{-4mm}
\begin{tabular}{c | c || c|c|c|c|c|c|}
    %\hline 
    \cline{2-8}% \textbf{\VGGFace2} 
    & $|\yg|$ ($N$) & 30 (10k) & 58 (20k) & 111 (40k) & 220 (80k) & 440 (160k) & 880 (320k) \\ \cline{2-8}
    & Random & $0.35 / 100$ & $0.67 / 100$ & $1.29  / 100$ & $2.55  / 100$ & $5.10 / 100$ & $10.2  / 100$ \\
    \hline
    \hline
     \multirow{2}{*}{\rotatebox{90}{\tiny \sc StyleGAN}} %\multirow{5}{c|}{~}
    & Threshold $T_{0}$  & $1.79 / 90$ & $3.08 / 77.6$ & $5.36 / 73.9$ & $7.56 / 58.6$ &  $\bf 9.25 / 38.9$ &  $\bf 12.5 / 30.6$  \\ \cline{2-8} 
    & Threshold $T_{1}$  &  $\bf 24.6 / 56.7$ &  $\bf 40.5 / 51.7$ & $\bf 69.5 / 36.9$ &  $\bf 75.5 / 16.8$  & $42.9 / 2.73$ & $14.3 / 0.341$ \\ \cline{2-8} 
    \hline
    \hline
     \multirow{2}{*}{\rotatebox{90}{\tiny \sc LSGAN}} %& Precision (in \%)
    & Threshold $T_{0}$  & $2.51 / 100 $ & $4.57 / 96.6 $ & $\bf 2.19 / 39.6$ & $\bf 3.37  / 28.6$ & $\bf 6.73 / 32.3$ & $11.9 / 28.5$ \\ \cline{2-8} 
    & Threshold $T_{1}$  & $\bf 49.1 / 93.3$ & $\bf 80.3 / 84.5 $ & $8.57 / 2.70 $ & $7.69 / 1.82$ & $16.7 / 0.68$ & $10.0 / 0.34$ \\ \cline{2-8} 
    \hline
\end{tabular}

\vspace{4mm}

\centering
\rotatebox[origin=c]{90}{\bf \CASIA}
\hspace*{2mm}
\renewcommand{\arraystretch}{1.5}\setlength{\tabcolsep}{5pt}
%\hspace*{-4mm}
\begin{tabular}{c | c || c|c|c|c|}
    %\hline 
    \cline{2-6}% \textbf{\VGGFace2} 
    & $|\yg|$ ($N$) & 30 (4.5k) & 58 (8.7k) & 111 (16.65k) & 220 (33k) \\ \cline{2-6}
    & Random & $2.32 / 100$ & $4.49 / 100$ & $8.59 / 100$ & $17.0 / 100$ \\ 
    \hline
    \hline
     \multirow{2}{*}{\rotatebox{90}{\tiny \sc StyleGAN}} %\multirow{5}{c|}{~}
    & Threshold $T_{0}$  & $7.16 / 80.0 $ & $13.7 / 79.3 $ & $ \bf 22.2 / 69.4 $ & $ 12.1 / 15.9 $  \\ \cline{2-6} 
     & Threshold $T_{1}$ & $ \bf 54.5 / 20.0 $ & $ \bf 90.0 / 15.5 $ & $ 81.8 / 8.11 $ & $ 9.09  / 4.55$ \\ 
    \cline{2-6} 
    \hline
    \hline
     \multirow{2}{*}{\rotatebox{90}{\tiny \sc LSGAN}} %& Precision (in \%)
    & Threshold $T_{0}$ & $ 9.54 / 83.3 $ & $ 16.0 / 75.9 $ & $ \bf 22.1 / 66.7 $ & $ \bf 34.5 / 53.2 $ \\ \cline{2-6} 
    & Threshold $T_{1}$ & $ \bf 93.3 / 46.7 $ & $ \bf 87.5 / 24.1 $ & $ 81.8 / 8.11 $ & $ 100 / 9.09$ \\ \cline{2-6} 
    \hline
\end{tabular}
\vspace*{3mm}
\caption{\textbf{Precision / Recall rate (in \%) for the Identity Membership Attack (Alg.~\ref{algo:attack}) on GANs trained on the \VGGFace2 and \CASIA datasets with low bias and varying diversity (setting 1 \S~\ref{sec:singlesplit}).}
\em 
 Higher is better, and corresponding best $F_{1}$ scores are emboldened.
 $N = \sum_{i \in \yg} N_{i}^G$ indicates the total number of individual samples used during training of the GAN and $|\yg|$ is the number of identities. 
 For the attack, $|\yf|$ identities are used ($|\yf|=8631$ for \VGGFace2, $|\yf|=1292$ for \CASIA).
 The baseline is given by \emph{random} guessing, which reaches a precision of $\tfrac{|\yg|}{|\yf|}$ whatever the recall. 
}
%\vspace*{1mm}
\label{table:membership attacks precision-recall \VGGFace2 and Casia}
\end{table}
%##################################################################

% ################ PR CURVES on VGGFACE2 with StyleGAN for Setting 1 #########################
%##################################################################

% ########### VGGFACE2 double splitting with Precision AND RECALL ###########
\begin{table}[htb] %[!p]% !htb
%\centerline{\em  Attack Precision / Recall (rate in \%) for setting 2 (\S~\ref{sec:doublesplit}): varying bias with high diversity}
%\vspace{2mm}

\centering
\rotatebox[origin=c]{90}{\bf \VGGFace2}
\hspace*{2mm}
\renewcommand{\arraystretch}{1.5}\setlength{\tabcolsep}{6pt}
%\hspace*{-4mm}
\begin{tabular}{c | c || c|c|c|c|}
    %\hline 
    \cline{2-6}% \textbf{\VGGFace2} 
    %& $|\ygone|$ ($N_1$), $|\ygtwo|$ ($N_2$)  &  &  &  &  \\
    & $|\ygone|$ ($N_1$) & 20 (6k) & 40 (12k)  & 80 (24k) & 160 (48k) \\ \cline{2-6} 
    &  $|\ygtwo|$ ($N_2$) & \multicolumn{4}{c|}{2000 (40k)}  \\ \cline{2-6} 
 
    & Random  & $0.23/100$ & $0.46/100$ & $0.93/100$ & $1.85/100$ \\
    \hline
    \hline
     \multirow{2}{*}{\rotatebox{90}{\tiny \sc StyleGAN}} %\multirow{5}{c|}{~}
    & Threshold $T_{0}$  & $0.70 / 75.0 $ & $1.25 / 65.0$ & $2.99 / 76.2$ & $3.83 / 49.4$ \\ \cline{2-6} 
    & Threshold $T_{1}$  & $\bf 33.3 / 30.0$ & $\bf 31.2 / 25$ & $\bf 56.2 / 22.5$ & $\bf 40 / 8.75$ \\ \cline{2-6} 
    \hline
    \hline
     \multirow{2}{*}{\rotatebox{90}{\tiny \sc LSGAN}} %& Precision (in \%)
    & Threshold $T_{0}$  & $0.46 /51.3 $ & $1.12 /54.8 $ & $2.33 / 62.5$ & $\bf 3.07 / 37.5$ \\ \cline{2-6} 
    & Threshold $T_{1}$  & $\bf 26.5 / 21.7$ & $\bf 29.8 /23.25 $ & $\bf 47.8 / 13.8$ & $5.56 / 18.8$ \\ \cline{2-6} 
    \hline
\end{tabular}
\vspace*{3mm}
\caption{\textbf{Precision / Recall rate (in \%) for the Identity Membership Attack (Alg.~\ref{algo:attack}) on GANs trained on the \VGGFace2 dataset in the second setting.}
\em 
 The dataset $\yg$ is now composed of two sets of distinct identities: $\ygone \cup \ygtwo$.
 GANs are trained on a large number of identities $|\yg| = |\ygone| + |\ygtwo|$ (exceeding 2000), with different bias towards a small set of identities (as measured by the ratio $|\ygone| / |\yg|$).
 $N_1$ and $N_2$ represents the number of samples for $\ygone$ and $\ygtwo$.
 For the attack, $|\yf=8631|$ identities are used (including $\yg$) and precision/recall are reported for identities from $\ygone$ against $\yf \cap \ygone$.
 The baseline is given by \emph{random} guessing, which is the proportion of training identities $\tfrac{|\ygone|}{|\yf|}$. 
}
%\vspace*{1mm}
\label{table:membership attacks precision \VGGFace2 setting2}
\end{table}
%##################################################################

% ################ PR CURVES on VGGFACE2 with StyleGAN for Setting 2 #########################
%##################################################################

\subsection{Analysis}

\begin{figure}[!htb]
    \centering
    \begin{tabular}{ccc}
    \raisebox{20mm}{\rotatebox[origin=c]{90}{\bf \VGGFace2}}
    &\includegraphics[width = .48\linewidth,trim={.4cm 0 1.5cm 0},clip]{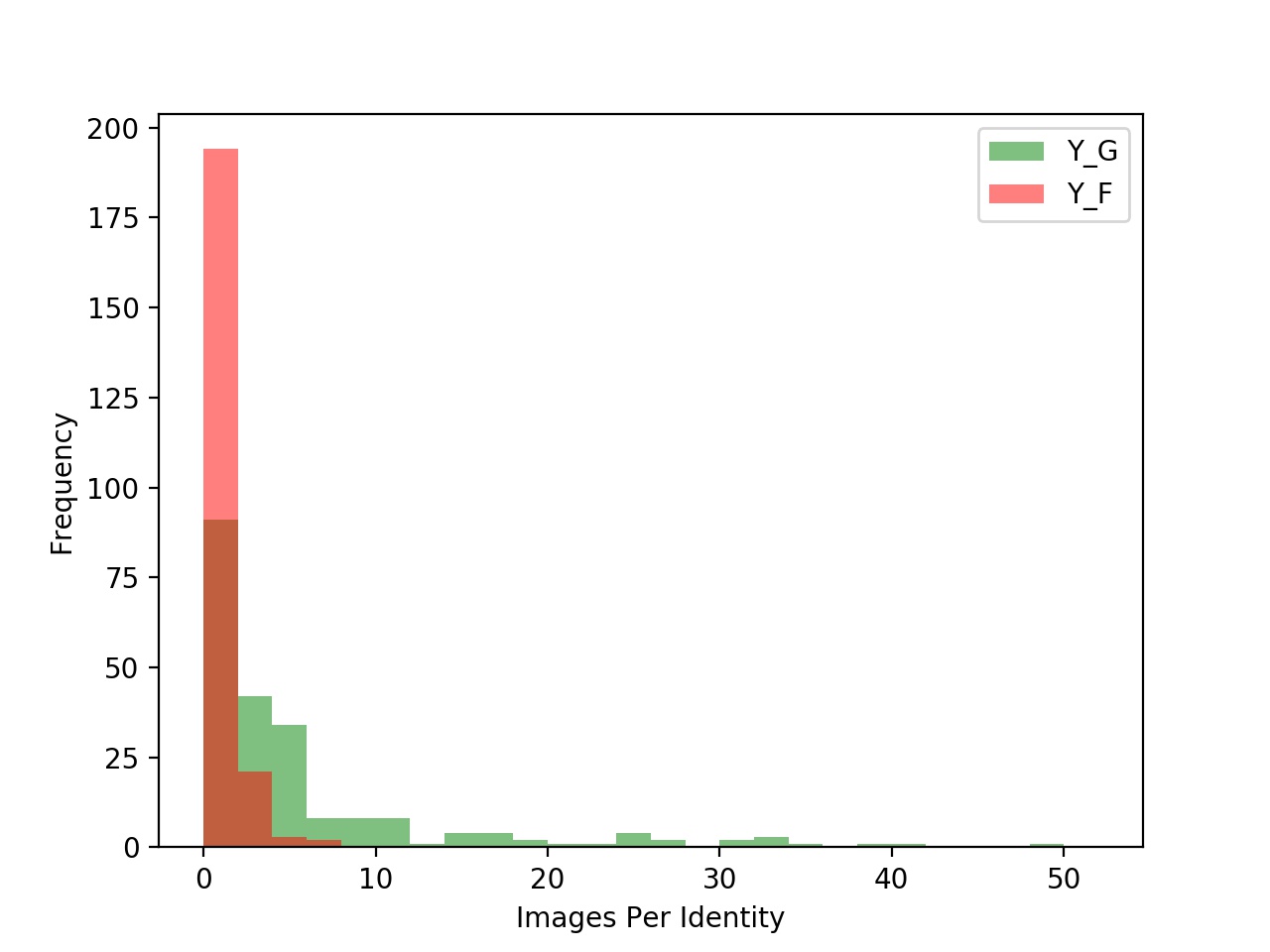}
    &\includegraphics[width = .48\linewidth,trim={.4cm 0 1.5cm 0},clip]{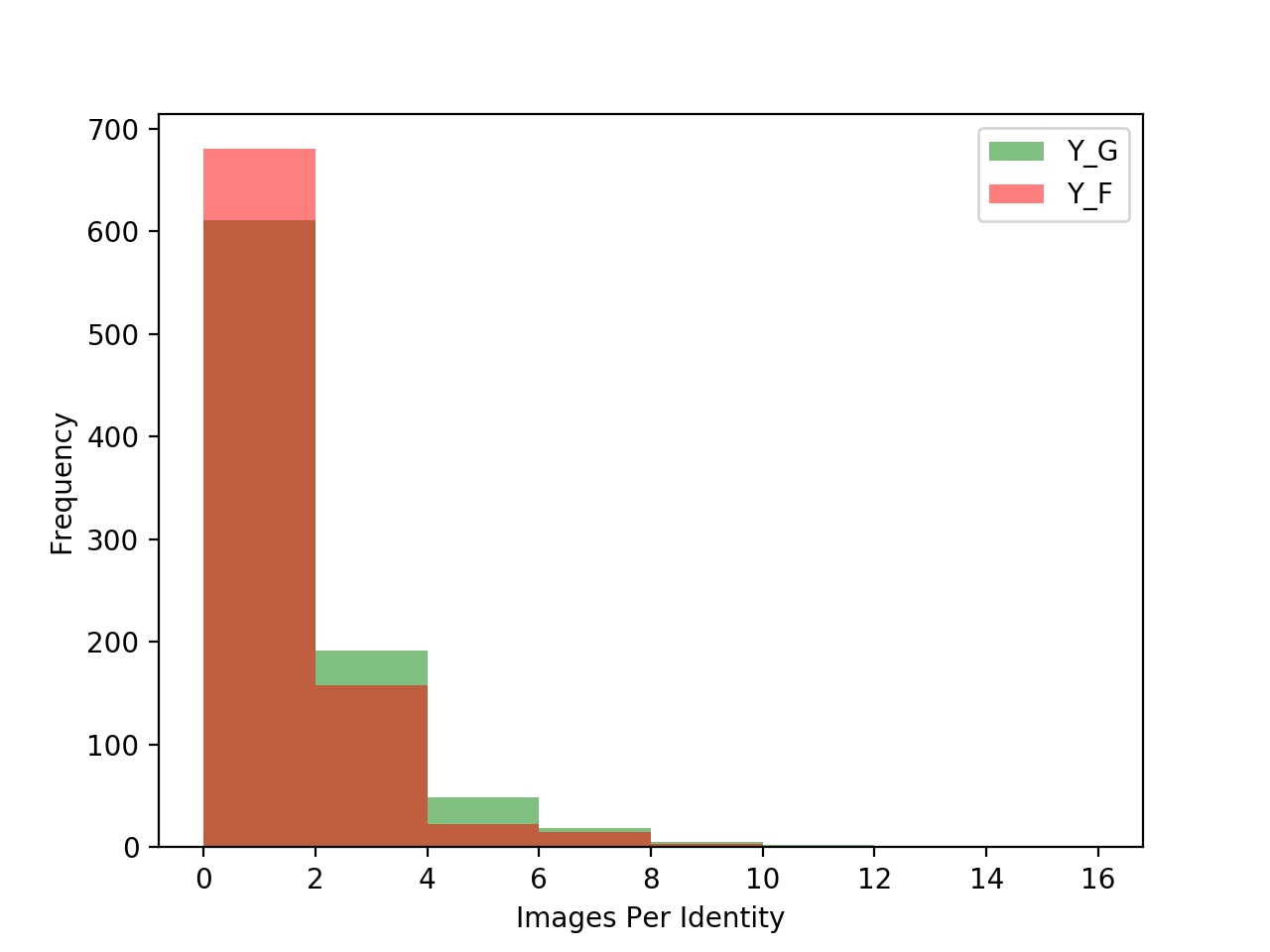}
    \\
    
    &$| \yg| = 220$, $N=66k$ 
    &$| \yg| = 880$, $N=264k$
    \end{tabular}
    
    \vspace{2mm}
    
    \centering
    \begin{tabular}{ccc}
    \raisebox{20mm}{\rotatebox[origin=c]{90}{\bf \VGGFace2}}
    &\includegraphics[width = .48\linewidth,trim={.4cm 0 1.5cm 0},clip]{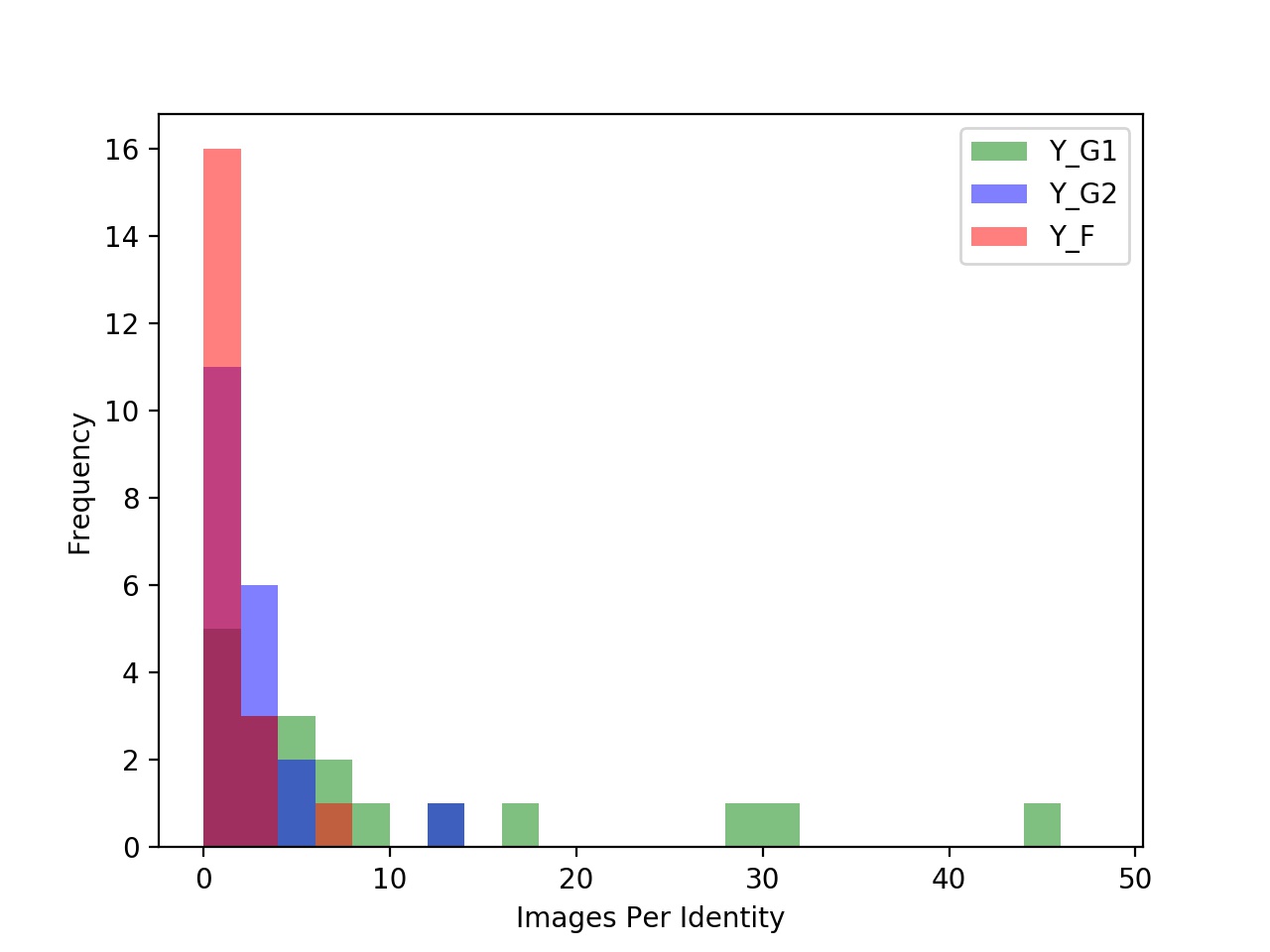}
    &\includegraphics[width = .48\linewidth,trim={.4cm 0 1.5cm 0},clip]{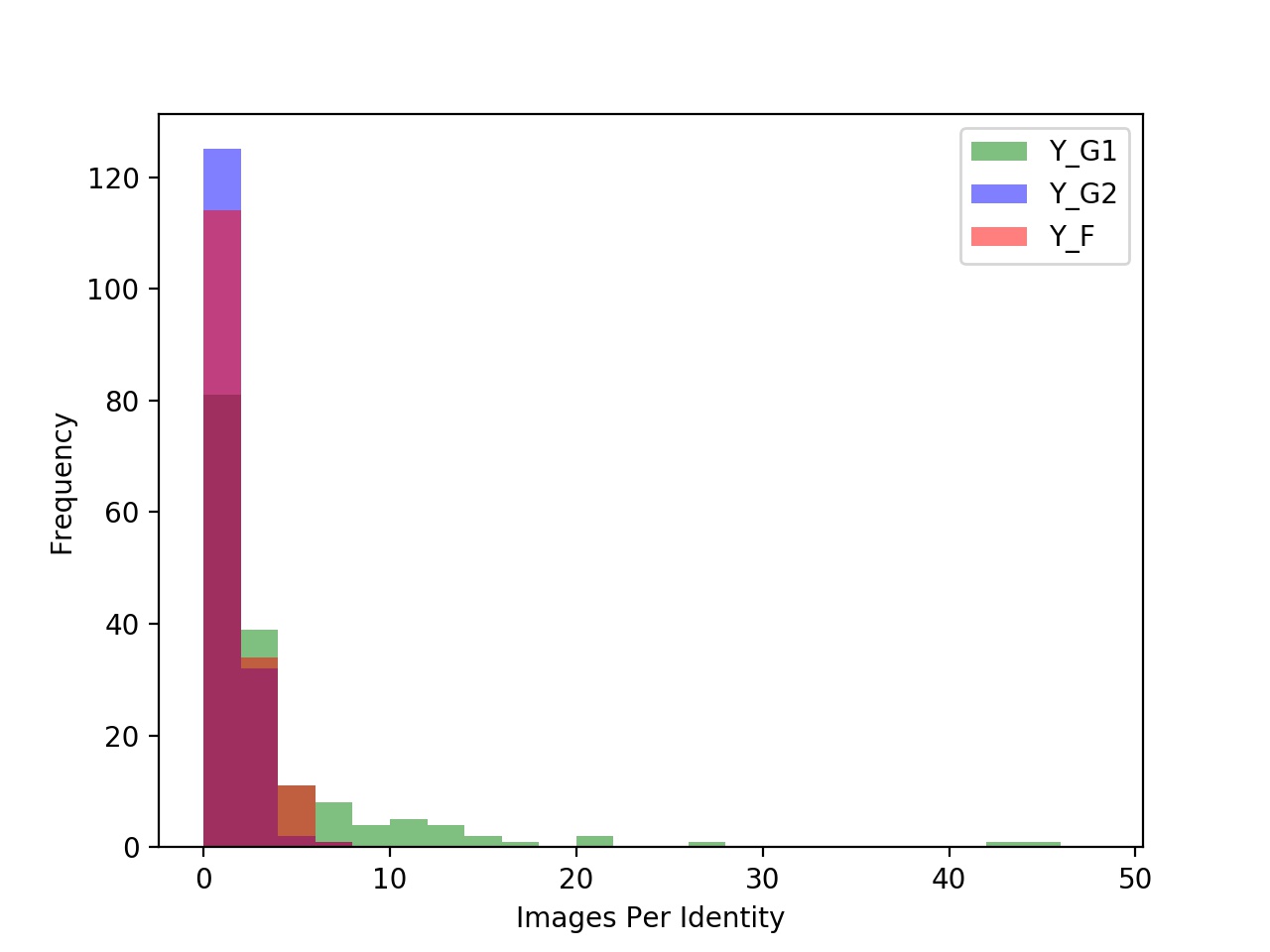}
    \\
    
    &$| \ygone| = 40$, $| \ygtwo| = 2000$,  $N=52k$ 
    &$| \ygone| = 160$, $| \ygtwo| = 2000$, $N=88k$
    \end{tabular}
    
    \caption{\textbf{Per identity frequency histograms of generated samples}.
     \em In the first row, StyleGAN generators are trained with unbiased data (setting 1), see Sec.~\ref{sec:singlesplit}. Without enough diversity, membership attacks are highly precise, as the bars in green (representing private training identities), can be easily distinguished. The second row showcases the high diversity \& high bias scenario in Sec.~\ref{sec:doublesplit}. Here, blue represents a third, diverse auxiliary set, which  cannot be distinguished, contrary to the biased samples (in green).
    }
    \label{fig:id_histograms_two_settings}
\end{figure}

Figure~\ref{fig:id_histograms_two_settings} shows the distribution of identities detected by the face identification network $F$ based on the number of samples recovered from StyleGAN trained on the \VGGFace2 dataset, for different settings.
Identities from $\yf \backslash \yg $ and $\yf$ are displayed in different colors to highlight the fact that the tail of the distributions corresponds to frequent recovered identities that are indeed in the training set $\yg$, and prone to membership attack. 
Observe that the distribution has a lighter tail when the number of identities $|\yg|$ and the number of samples $N$ increase (first scenario), or when the bias ratio $\frac{|\ygone|}{|\yf|}$ decreases (second scenario).
As discussed later in Section~\ref{sec:visual_evaluation}, samples from these identities can be extracted to visually assess overfitting, as done in Fig.~\ref{fig:teaser}.

In Table~\ref{table:membership attacks precision-recall \VGGFace2 and Casia}, it is clear that insufficient diversity yields GANs models that are susceptible to membership attacks, even in the blind setting. With enough diversity (here, towards 880 identities), attacks are reduced to near guessing. 
Table~\ref{table:membership attacks precision \VGGFace2 setting2}
and Figure~\ref{fig:PR_curve_biased}
show that diversity %in and of 
itself is not enough to protect the data in the presence of bias. Even after taking more than 2000 identities, attacks still succeed with high $F_{1}$ scores. 
As for the difference across GAN models, LSGAN seems to be far more susceptible to attacks until a certain data set size (in Table~\ref{table:membership attacks precision-recall \VGGFace2 and Casia}, at $\yg=111$ for \VGGFace2), where attack success significantly drops off. We assume that this is because the LSGAN model we trained had nearly 10x less parameters than the StyleGAN model and likely could not overfit samples after some amount of diversity. On the other hand, StyleGAN models are far higher quality. We leave investigation of the trade off between model size, quality and susceptibility to attack to future work. 

\subsection{Early Stopping}

\begin{figure}[!htb]
    \centering
    \begin{tabular}{cc}
    % trim=left bottom right top, clip
        \includegraphics[width=.49\linewidth, trim=85 10 100 90, clip]{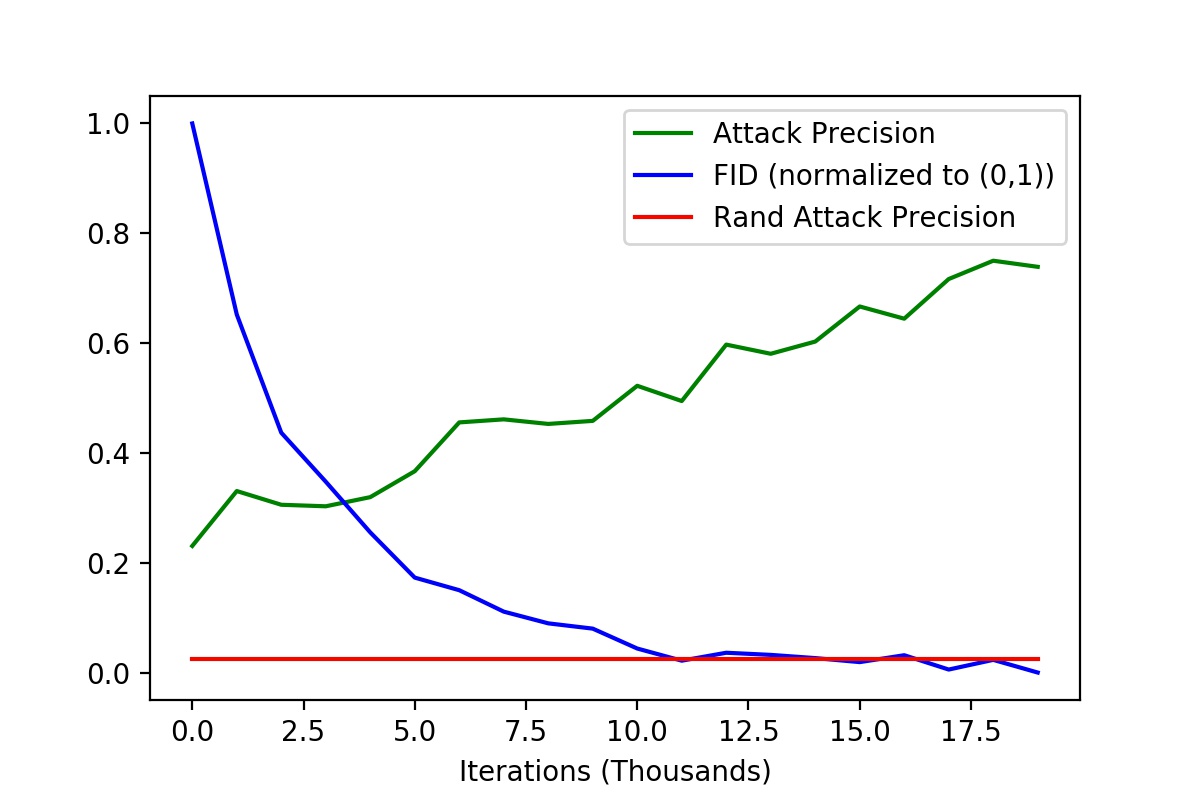}
        &\includegraphics[width=.49\linewidth, trim=85 10 100 90, clip]{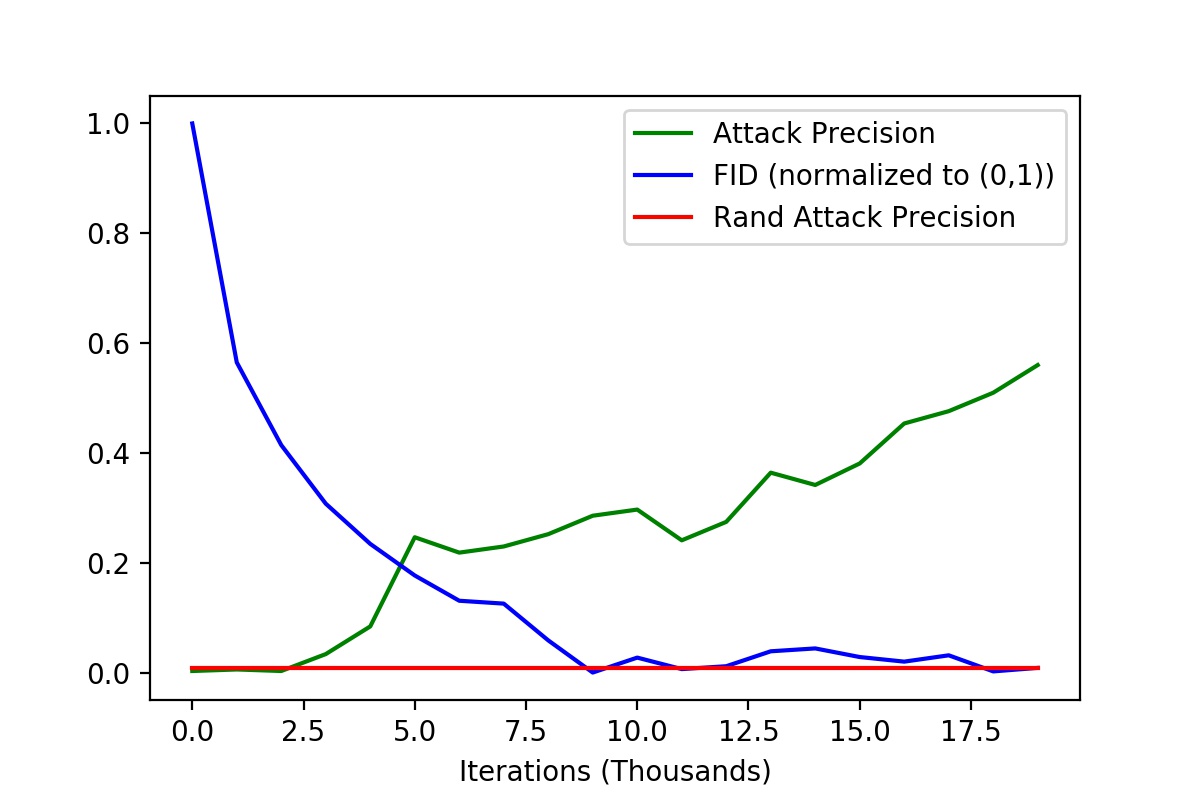}
        \\[2mm]
        Setting 1 ($|\yg|=220$)
        & 
        Setting 2 ($|\ygone|=20$ \& $|\ygtwo|=2000$)
    \end{tabular}
    
    \caption{\textbf{FID  vs attack precision.} 
    \em  The graphs above suggest that early stopping may be useful in GAN training, w.r.t. privacy. While the FID converges, membership attacks continue to gain precision. }
    \label{fig:early_stopping}
\end{figure}

Thus far, we have only explored dataset size and diversity as possible factors effecting attack efficacy. However, a common tool in image classification to prevent overfitting is early stopping. In order to asses GAN sample quality, we refer to the commonly used Frechet Inception Distance (FID) \cite{heusel2017gans}. Fig.~\ref{fig:early_stopping} investigates the interaction between FID and the number of GAN training iterations. Indeed, beyond a certain iteration (here for StyleGAN, around 17000), continuing to train is detrimental to privacy, with no gain in FID.

%\clearpage

\section{Discussion and Conclusion}\label{sec:conclusion}

In this work, we exposed several properties of GANs trained on facial data not previously discussed in the literature. By controlling the identities seen during training and subsequently detecting those identities with a face identification network, we demonstrated a successful blind membership attack. We identified several factors influencing susceptibility to attack, most notably that datasets with more identities have less detectable overfitting. Furthermore, dataset diversity alone will not protect against the presence of dataset bias. Finally, more iterations seemed to exacerbate attack success while not necessarily better image quality. 
\paragraph{}
Contrary to most membership attack in the literature, this is a pure \emph{black-box} attack in the sense that it does not require any further training based upon the generator, nor additional information about its architecture and parameters \cite{sablayrolles2019white}.
In addition, our attack is driven directly against the generator, and it did not require the exact training samples to be successful, as in the LOGAN approach.

Yet, the proposed approach is still restricted in various ways.
To begin with, privacy issues is not limited to face datasets. Generalizing this work 
to other modalities (such as other biometric data, medical images, or other media) that are scrutinized with specialized networks is crucial. 
% other  applications (\emph{e.g.} Person Re-Identification).
Furthermore, entirely ruling out overfitting is much more challenging than detecting it.
Indeed, as we showed that the proposed attack is less effective as the number of identities grows, 
%ruling out that a given identity is not reproduced is more difficult. 
it strongly depends on the performance of the identification network.
More effort needs thus to be put to improve detection. A first direction would be to improve the attack itself by designing more sophisticated mechanisms. For instance, by training the identification network specifically for this task (to measure the generalisation upon the discovery of new identities rather than a fixed number of ones from the training set). We also note that while the attack doesn't require training samples, it does require $\yg$ and $\yf$ to have some amount of overlap. Clearly, the attack precision would be zero if $\yf \cap \yg =\emptyset$. 
%Furthermore, confidence about how accurate your attack is relates to your confidence about how large this intersection is, as we used this intersection size to evaluate the "random guess" attack. 

%\paragraph{}
Finally, this work addresses a problem with training GANs on sensitive data (e.g. copyrighted or private face images). Careful dataset curation or early stopping may mitigate these problems. On the other hand, more sophisticated solutions may exist, such as those integrated into the GAN objective function, or those which directly modify the generation, such as distillation, to promote privacy. We leave this for future work.

For the time being, ensuring sufficient face dataset diversity (in terms of number of identities), low dataset bias and early stopping (when the FID stops changing, or reaches a threshold) all appear to be good empirical tools to train safer models.

\clearpage
\bibliographystyle{splncs}
\bibliography{bib}

%\clearpage
%{\small
%\bibliographystyle{ieee_fullname}
%\bibliographystyle{splncs04}
%\bibliography{bib}
%\printbibliography
%}
\end{document}